\documentclass{article}

   \usepackage[dblblindworkshop, final]{neurips_2026}
\usepackage{comment}
\workshoptitle{The 6th Workshop on Mathematical Reasoning and AI}

\usepackage[utf8]{inputenc} 
\usepackage[T1]{fontenc}    
\usepackage{hyperref}       
\usepackage{url}            
\usepackage{booktabs}       
\usepackage{amsfonts, amsmath, amssymb, amsthm}  
\usepackage{thmtools}
\usepackage{nicefrac}       
\usepackage{microtype}      
\usepackage{xcolor}         
\usepackage{graphicx}
\usepackage{subcaption}

\newtheorem{proposition}{Proposition}
\newtheorem{lemma}{Lemma}

\title{Can Small Transformers Learn the Language of Coxeter Groups?}

\author{Elisenda Grigsby\thanks{Corresponding author. Authors listed in alphabetical order.
    } \\
  Department of Mathematics\\
  Boston College\\
  Chestnut Hill, MA 02467 \\
  \texttt{grigsbyj@bc.edu} \\
  \And
  Yiyao Huang \\
  Department of Mathematics\\
  Boston College\\
  Chestnut Hill, MA 02467 \\
  \texttt{huangaxo@bc.edu}
   \And
  Arushi Katyal \\
  Department of Mathematics \\ 
  Boston College \\
  Chestnut Hill, MA 02467 \\
  katyala@bc.edu \\
   \And
  Ryan Lin \\
  Department of Mathematics\\
  Boston College\\
  Chestnut Hill, MA 02467 \\
  \texttt{linrya@bc.edu} \\
   \And    
  Pranavkrishnan Ramakrishnan \\
  Department of Mathematics\\
  Boston College\\
  Chestnut Hill, MA 02467 \\
  \texttt{ramakrip@bc.edu} \\
  \And
  Weiya Wang \\
  Department of Mathematics\\
  Boston College\\
  Chestnut Hill, MA 02467 \\
  \texttt{wangfbv@bc.edu}
}

\begin{document}

\maketitle

\begin{abstract}
  The field of mechanistic interpretability is well-suited to collaborative projects between mathematicians and AI researchers. This work describes one such project. We trained a small transformer to learn the language of the Coxeter group $\widetilde{A}_2$, best known as the symmetries of the tiling of the Euclidean plane by equilateral triangles. We then formulated and proved a word-theoretic conjecture about $\widetilde{A}_2$ after observing the attention patterns of the trained model.
\end{abstract}

\section{Introduction}
State of the art chain of thought (CoT) reasoning models, built on large language models (LLMs),  trained on the entire corpus of human mathematical knowledge, and benchmarked by mathematicians, have achieved - and will likely continue to achieve - impressive performance on frontier-level mathematics, cf. \citet{GeorgievGomezSerranoTaoWagner2025, AlexeevBarretoLiLichtmanPriceShahTangTao2026, AlonBloomGowersLittSawinShankarTsimermanWangWood2026}. Most such models were designed and trained by researchers at AI companies who have access to staggering levels of funding - and hence computational resources - that are out of reach for even the best-funded researchers in academia, \citet{Hao2025}. Although more recent models are integrated with formal proof verification systems like Lean that can in principle autonomously certify the correctness of a mathematical result, it can be challenging for humans to verify that autoformalized concepts match the authors' intent \citet{JiangStaatsSzegedyRabeJamnikLiWu2022, LiWuLiWeiZhangYangMa2024, zhang-etal-2025-autoformalization}. The end result is that producing new mathematics is not truly a collaboration between humans and AI systems, and mathematicians using these systems understand neither the structure of the models they are using nor how they arrive at the results that they do.

In contrast, the aim of {\em mechanistic interpretability} research \citet{anthropic2022transformer}, \citet{nanda2022comprehensive}, \citet{SharkeyEtAl2025} is to develop a concrete understanding of the inner workings of AI models, especially the transformer architectures that have had the largest recent impact on the AI landscape. Applying mechanistic interpretability techniques to mathematical problems that are either geometric or combinatorial in nature is a promising avenue for mathematicians and AI researchers to truly collaborate.  

The field of {\em geometric group theory} (cf. \citet{Bridson1998GeometryWordProblem}), whose objects can frequently be understood via the mathematical language of {\em group presentations}, is a fertile playground for such collaborations. A (finite) group presentation is a way of encoding individual symmetries (elements of a {\em group}) as strings (words) in a finite set of letters, up to a prescribed finite set of equivalences. In addition, many groups are {\em automatic} --there exist finite state automata that can perform word processing in the group \citet{EpsteinCannonHoltLevyPatersonThurston1992}. This suggests, cf. \citet{LiuAshGoeelKrishZhang}, that using AI models like transformers to study word-theoretic questions in these groups can benefit both the field of mechanistic interpretability and the field of geometric group theory.

\begin{figure}
    \centering    \includegraphics[width=0.4\linewidth]{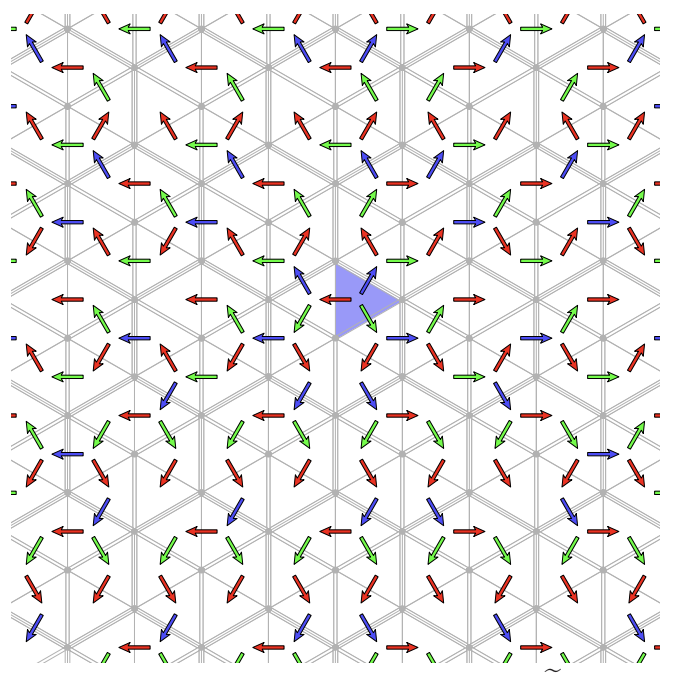}
    \caption{ShortLex words in $\widetilde{A}_2$, visualized as paths of reflections in the $\mathbb{R}^2$--tiling on which it acts}
    \label{fig:A2tilde}
\end{figure}

\begin{figure}[h]
    \centering        \includegraphics[width=0.6\linewidth]{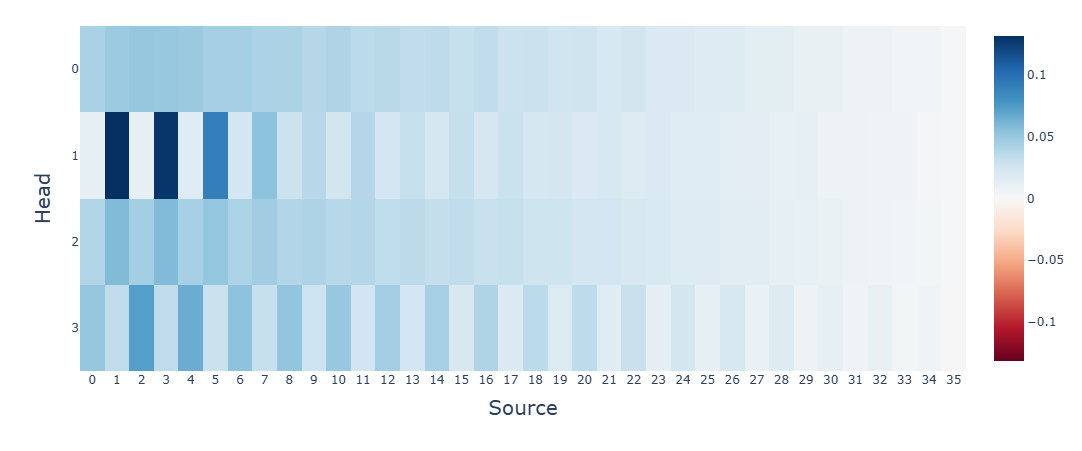}
    \caption{Right descent set prediction task: average attention paid by the final token in each of the $4$ attention heads, averaged over ShortLex normal form words of length $36$ in $\widetilde{A}_2$.} 
\label{fig:RightDescents}
\end{figure}

\begin{figure}[h]
    \centering
    
        \centering    \includegraphics[width=0.6\linewidth]{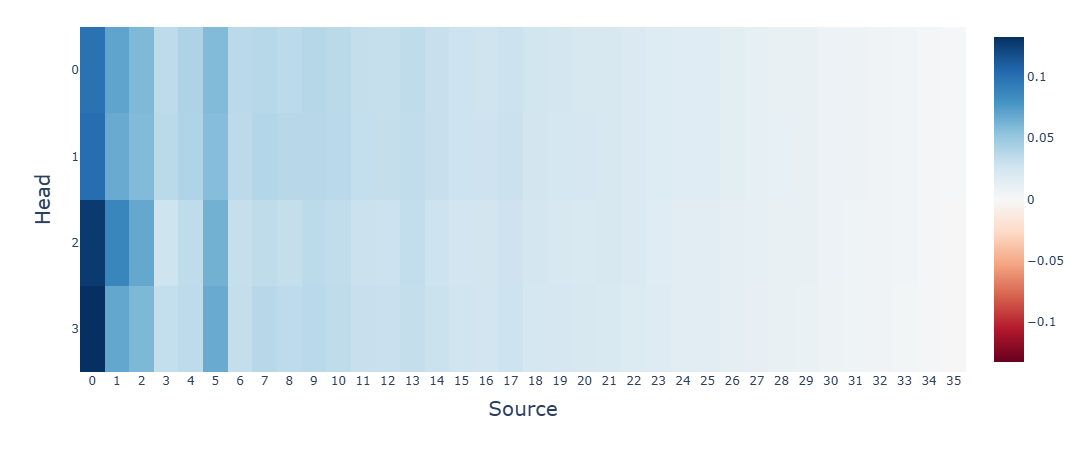}
        \caption{Left descent prediction: Average attention of final token of length $36$ normal form word}
\label{fig:LeftDescents}
\end{figure}

In the present work, we describe two (small) geometric group theory results (Propositions \ref{prop:OddIndicesDetermineLeftDescent} and \ref{prop:6LetterEnough}) we proved after observing the attention patterns (Figures \ref{fig:RightDescents} and \ref{fig:LeftDescents}) of a (small) transformer model trained on a synthetic data set for the automatic Coxeter group $\widetilde{A}_2$. $\widetilde{A}_2$ is the (infinite) group of symmetries of the tiling of the Euclidean plane by equilateral triangles, see Figure \ref{fig:A2tilde} -  taken from \citet{casselman2008computationcoxetergroupsii} - and the amazing on-line visualization tool \citet{Gibson2022Lievis}. It is well-known that:
\begin{itemize}
    \item $\widetilde{A}_2$ is {\em generated} by the $3$ reflections $a_1, a_2, a_3$ in the hyperplanes adjacent to the purple triangle in Figure \ref{fig:A2tilde}; i.e., {\em any} symmetry can be realized as a word in these three letters, and
    \item each pair $a_i \neq a_j$ of these reflections is {\em braid-flippable}, i.e., satisfies the simple relation that $a_ia_ja_i \sim a_ja_ia_j$ define the same symmetry; equivalently, the string $(a_ia_j)^3$ of reflections does nothing - which can be seen from Figure \ref{fig:A2tilde} by noting that $a_ia_j$ rotates the figure $120$ degrees about the corresponding vertex of the purple triangle.
\end{itemize}

We trained a $256$D transformer, with a single $4$--head attention layer and an MLP layer of dimension $1024$
to learn {\em left} and {\em right} descent sets of length $\leq 36$ group elements in $\widetilde{A}_2$. Left (resp., right) descents of a Coxeter group element $g$ are the possible first (resp., last) letters of minimal-length words representing  $g$. Note that {\em minimal length} words in $\widetilde{A}_2$ are not unique, but their {\em ShortLex normal form} expressions are. The ShortLex normal form of a Coxeter group element is the unique minimal-length word that is lexicographically first with respect to a chosen ordering on the generators. 

\begin{figure}[h]
    \centering
    \begin{subfigure}{210 pt}
    \centering
    \includegraphics[width=\linewidth]{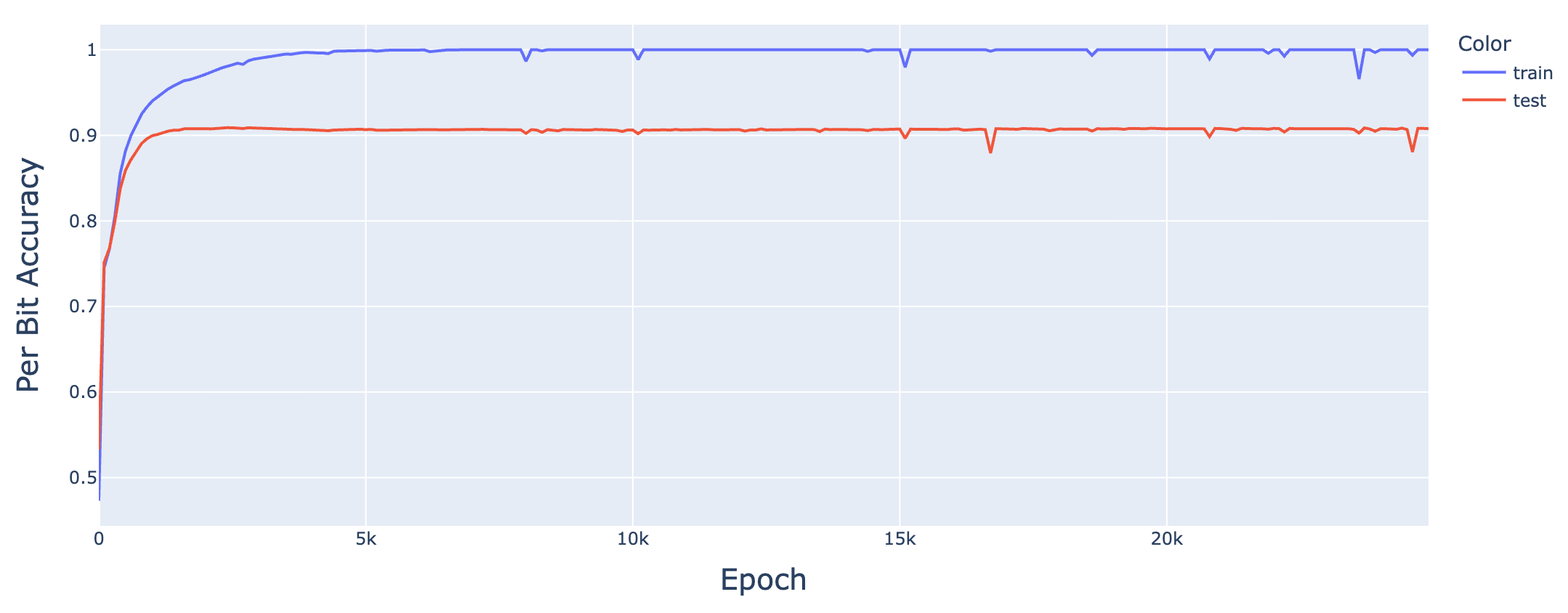}
    \caption{Model accuracy for random words of length $10$}  \label{fig:RandomAccuracy10}    
    \end{subfigure} \hfill
    \begin{subfigure}{185 pt}
    \centering  \includegraphics[width=\linewidth]{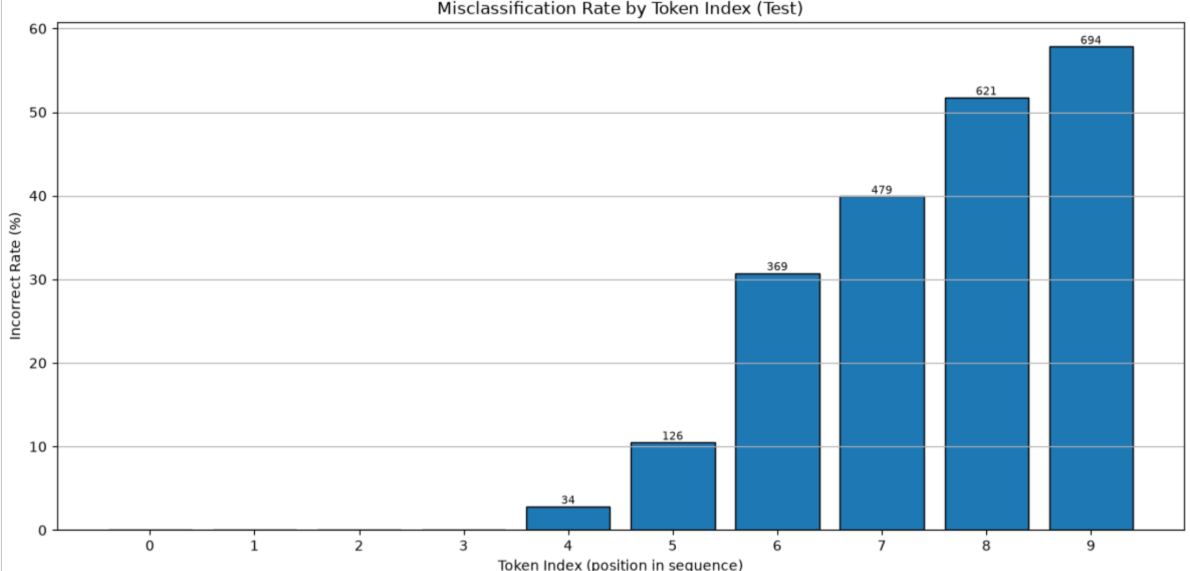}
    \caption{Misclassification by index for random words.}  \label{fig:RandomTestMisclass}
\end{subfigure}
\caption{Model performance for the left descent task on random words of length $10$ in $\widetilde{A}_2$.}
\label{fig:RandomModelPerformance}
\end{figure}

\begin{figure}[h]
    \centering
    \begin{subfigure}{210 pt}
    \centering
    \includegraphics[width=\linewidth]{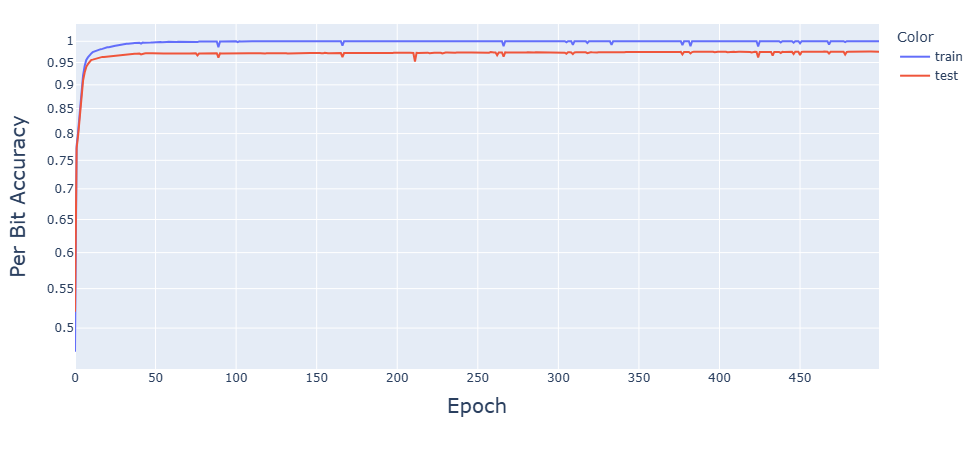}
    \caption{Model accuracy for reduced  words of length $10$}  \label{fig:ReducedAccuracy10}    
    \end{subfigure} \hfill
    \begin{subfigure}{185 pt}
    \centering    \includegraphics[width=\linewidth]{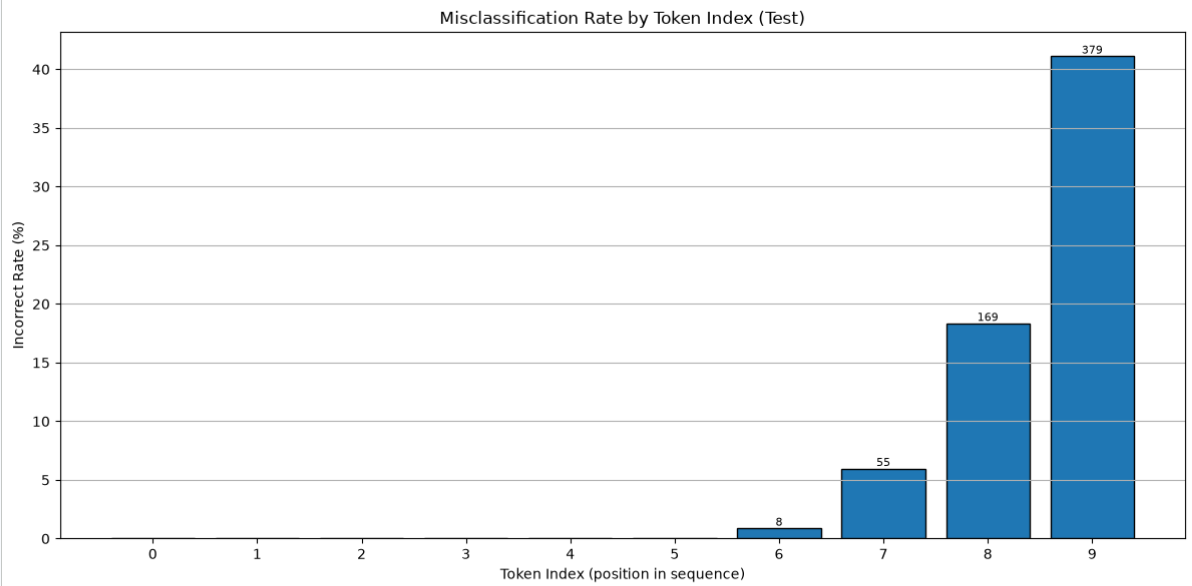}
    \caption{Misclassifications by index for reduced words.}  \label{fig:ReducedTestMisclass}
\end{subfigure}
\caption{Model performance for the left descent task on reduced words of length $10$ in $\widetilde{A}_2$}
\label{fig:ReducedModelPerformance}
\end{figure}

\begin{figure}[h]
    \centering
    \begin{subfigure}{195 pt}
    \centering    \includegraphics[width=\linewidth]{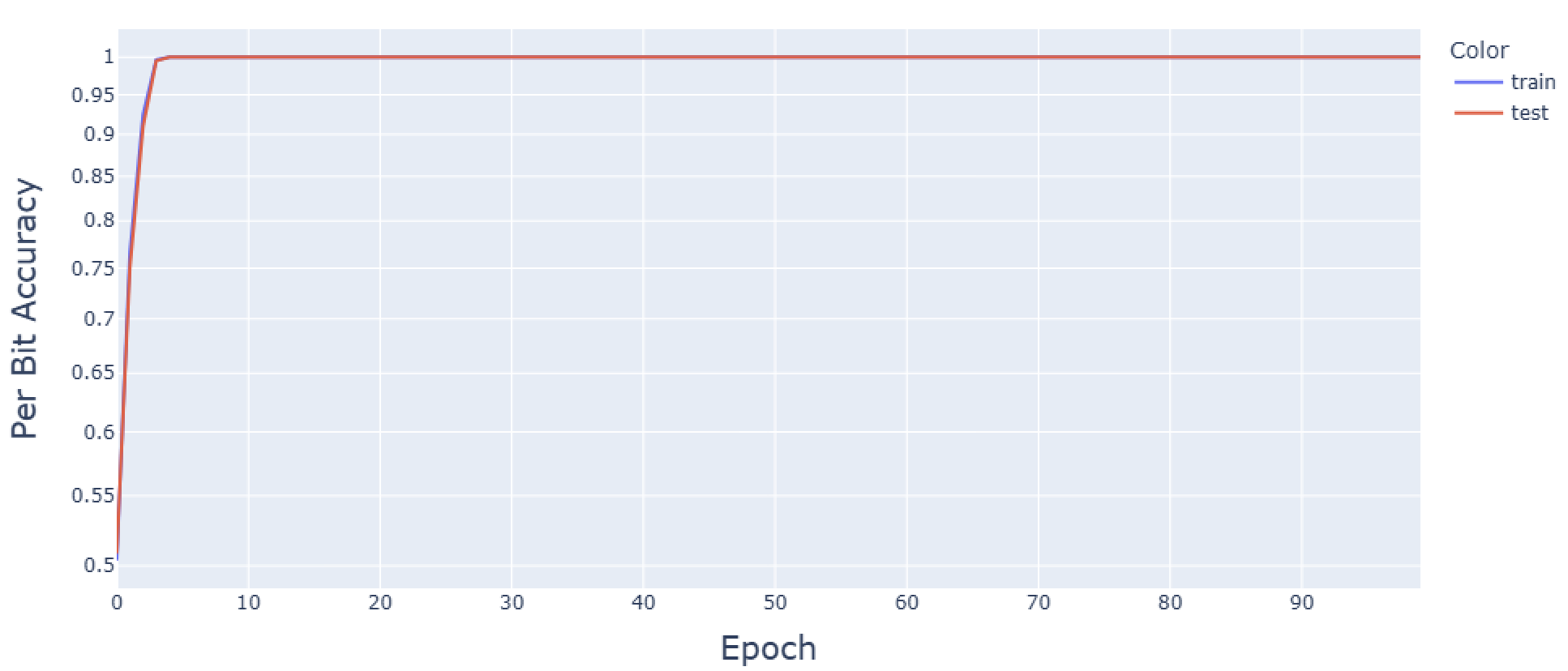}
    \caption{Model accuracy for ShortLex words of length $\leq 36$}  \label{fig:LeftDescentAccuracy}
\end{subfigure}
    \begin{subfigure}{195 pt}
    \centering    \includegraphics[width=\linewidth]{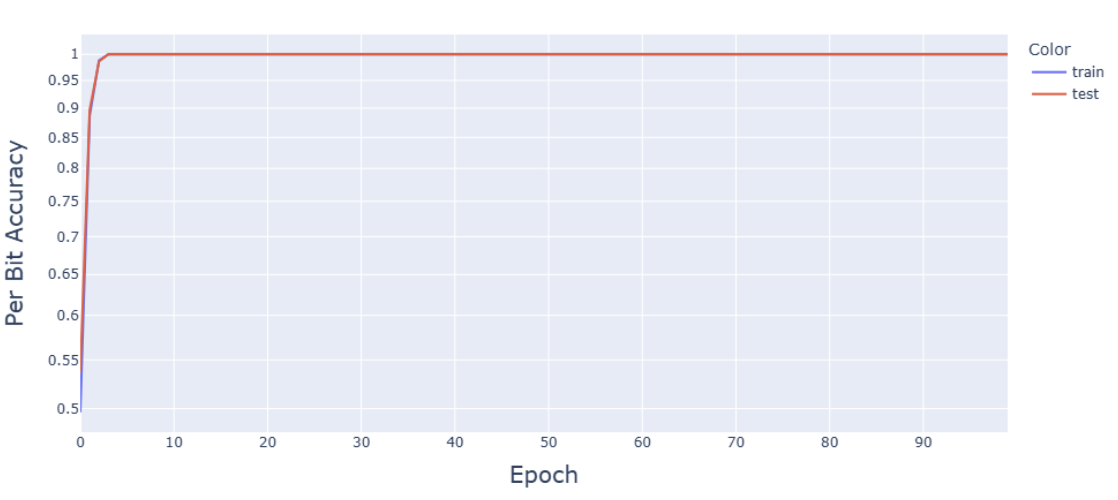}
    \caption{Model accuracy when all but first $5$ tokens masked}  \label{fig:AllButFirst5MaskedAccuracy}
\end{subfigure}
\caption{Model accuracy for the left descent task on ShortLex words in $\widetilde{A}_2$ of length up to $36$. The model achieved nearly $100 \%$ test accuracy (left), and performance remained strong (right) when all but the first $5$ tokens were masked, supporting Proposition \ref{prop:6LetterEnough}.}
\label{fig:ShortLexModelPerformance}
\end{figure}

We used three different data sets of increasing quality (unreduced, reduced, and ShortLex words representing elements of $\widetilde{A}_2$) in our experiments and found an accompanying clear improvement in model performance (Figures \ref{fig:RandomModelPerformance}, \ref{fig:ReducedModelPerformance}, and \ref{fig:ShortLexModelPerformance}) - test accuracy began deteriorating at length $5$ for the unreduced data set, at length $8$ for the reduced data set, and never for the ShortLex data set, which contained all ShortLex words of length 36. Our key takeaways are:

\begin{itemize}
    \item The attention patterns for the right descent task (Figure \ref{fig:RightDescents}) exhibit a same--parity structure, which motivated Proposition \ref{prop:OddIndicesDetermineLeftDescent}.
    \item The attention patterns for the left descent task (Figure \ref{fig:LeftDescents}) concentrate strongly on the beginning of the word, which motivated Proposition \ref{prop:6LetterEnough}.
    \item Quality and consistency of the data set was the most important factor determining model performance. Compare Figures \ref{fig:RandomModelPerformance} and \ref{fig:ReducedModelPerformance} with \ref{fig:ShortLexModelPerformance}.
 
\end{itemize}

\section{Related Work}
Mechanistic interpretability (cf. \citet{anthropic2022transformer, nanda2022comprehensive, Conmy, olsson2022induction}) is concerned with analyzing how the internal behavior of a model's parameters and latent features influences its outputs. Algorithmic tasks are a good fit, since a proposed circuit can be checked against a previously-known algorithm. The works that most directly influenced ours are \citet{nanda2022grokking}and \citet{Zhong}, which studied an (abelian) modular addition task and found evidence for Fourier-type circuits, and \citet{chughtai2023toy}, which studied a group multiplication task in some non-abelian finite groups, proposing a circuit implementing a representation-theoretic algorithm. In \citet{stander2024grokking}, the authors revisited the interpretation given in \citet{chughtai2023toy}, proposing an alternate interpretation in terms of quotient groups -- a reminder that even among experts, there is disagreement about what a model might have learned. 
In all of these works, the solution itself was well-known, and the mechanistic interpretability work provided evidence that the model learned the proposed algorithm.  Our work differs mainly in the complexity of the mathematical task. All of the works above explored a fixed operation on a {\em finite} group. We investigate an infinite group that enjoys a known finite word-processing algorithm \citet{Brink1993}. In our case, although an algorithm was known, the two results suggested by the model's solution - that we then proved without the aid of the model - do not follow immediately from the algorithm. We were inspired by \citet{LiuAshGoeelKrishZhang}, whose authors proposed a theoretical framework for understanding why transformer architectures might be particularly well-suited for learning elementary computational models called {\em finite state automata} or {\em finite state machines}. As far as we know, no prior mechanistic interpretability work has studied normal form expressions in infinite groups as a learning target.

\section{Mathematical background}\label{sec:math-background}

We save a more detailed discussion of groups, group presentations, and general Coxeter groups for Appendix \ref{app:GroupsandCoxeter}. For now, suffice it to describe our main object of interest, the \textit{Coxeter group of type $\widetilde{A}_2$}:
\begin{equation}
\label{eq:A2tildepresentation}
\widetilde{A}_2 = \langle a_1, a_2, a_3 : a_i^2 \textup{ for all } i, \,\, (a_ia_j)^3 \textup{ for all } i\neq j\rangle.
\end{equation}
Elements of $\widetilde{A}_2$ can be represented by words (finite ordered strings) with letters in the set $S = \{a_1, a_2, a_3\}$ of {\em generators}. Moreover, two words $w,w'$ represent the same group element if $w'$ can be obtained from $w$ by finitely many insertions or deletions of the {\em relations}, which are the strings to the right of the colon in the {\em group presentation} for $\widetilde{A}_2$ given in  Equation \ref{eq:A2tildepresentation}.

Recalling that $\widetilde{A}_2$ can be understood as the symmetries of the tiling in Figure \ref{fig:A2tilde}, the relations $a_i^2$ arise because the generators are reflections, and the relations $(a_ia_j)^3$ (called {\em braid}  relations) arise because the product of two reflections across hyperplanes that intersect in an angle of $\theta$ produce a rotation of $2\theta$. Each group element $g$ can be represented by infinitely many different words; Let  $\ell(g)$ denote the length of a shortest word in the $a_i$ representing $g$. From this, we may define the \textit{left descent set} $D_L(g)$ and \textit{right descent set} $D_R(g)$ of $g$:
$$
D_L(g) = \{s\in S: \ell(sg) < \ell(g) \},
$$
$$
D_R(g) = \{s\in S: \ell(gs) < \ell(g) \}.
$$
Given a word $w=s_1\ldots s_k$ with $s_i\in S$, we say it is of \textit{minimal length} if $\ell(w)=k$. We see that an element of $\widetilde{A}_n$ is not represented by a unique word of minimal length. For example, it follows from the relations that $a_1a_2a_1$ and $a_2a_1a_2$ are two minimal-length words representing the same group element in $\widetilde{A}_2$. To pick a distinguished minimal length representative, we pick the one that is first with respect to the lexicographic order $a_1 < a_2 < a_3.$ We say a word $w$ is in \textit{(ShortLex) normal form} if it is the unique word representing a group element that is first in lexicographic order. In the example above, $a_1a_2a_1 < a_2a_1a_2$, and therefore $a_1a_2a_1$ is the ShortLex normal form representative. In addition, we will say a word is \textit{reduced} if it contains no sub-word of the form $a_ia_i$ for any $i$.

We formulated and proved the following two results after observing the attention patterns (Figures \ref{fig:RightDescents} and \ref{fig:LeftDescents}) in the transformer model trained on the ShortLex data set. The statements are likely to be known by experts, but we never found explicit references to them in the literature. Their proofs are in Appendix \ref{app:GroupsandCoxeter}.

\begin{proposition}[Odd-indices determine right descent]\label{prop:OddIndicesDetermineLeftDescent}
Let $w$ be a length $n \geq 2$ word which is minimal length for an element of  $\widetilde{A}_2$. The right descent set of $w$  is a subset of the final two letters of $w$, contains the final letter of $w$, and
contains the penultimate letter of $w$ iff $w$ contains a braid-flippable pattern at an index $i$ with $i \equiv n \mod 2$. I.e., there is some $ i \equiv n \mod 2$ with $s_i = s_{i+2}$.
\end{proposition}

\begin{proposition}[Bounded initial string determines left descent]\label{prop:6LetterEnough}
The left descent set of any element of $\widetilde{A}_2$ is completely determined by the first six letters of its ShortLex normal form representative.
\end{proposition}

\section{Experimental setup and results}

We created three different increasingly structured $\widetilde{A}_2$ data sets for training: random words, reduced words, and ShortLex normal form words. For the random-word experiment, we generated 2,000 words of length 10 at random from $S = \{a_1, a_2, a_3\}$, allowing repetition of adjacent generators. The reduced-word dataset similarly contained 2,000 words of length 10, but adjacent generators were required to be different. For both datasets, we used a $40/60$ train/test split. Because there are substantially fewer ShortLex normal form words of any given fixed length, and we wanted to ensure that no complete test words appeared in the training set, we used all 1,998 ShortLex normal-form words of length at most 36, with a 30/70 train--test split. 
Words were labeled by either their left or right descents at each prefix. Labels were binary $3$--tuples, with a $1$ indicating membership of the corresponding generator in the descent set. For example, the label $(1,0,1)$ at position $1\leq k \leq n$ in a word $w = s_1 \cdots s_k \cdots s_n$ would indicate that there exist minimal length words representing the group element $s_1 \cdots s_k$ ending in $a_1$ and $a_3$, but not $a_2$.

Across all experiments, we built atop the \textbf{TransformerLens} \citep{nanda2022transformerlens} package, an open-source library for mechanistic interpretability of GPT-2-style models. For all experiments we used a $256$--dimensional model with a single attention layer with $4$ heads and a hidden MLP layer of dimension $1024$. 
The output layer was dimension $3$ with logistic sigmoid activation on each output neuron to predict membership in the descent set. The model was trained autoregressively with component-wise binary cross entropy loss, with a weight decay of $0.5$.

In the ShortLex experiments, the model generalized very well to data it hadn't seen. Of all test subwords from our ShortLex dataset on the left descent task, $2\%$ were unseen during training, and all were of length $>26$. The model achieved $100 \%$ accuracy on them. A second run with a  $90/10$ train test split achieved $99.3 \%$ accuracy on unseen subwords, with $12.4 \%$ ($5462$ subwords) unseen during training. Finally, on a manually split dataset consisting of all 63 words of length $\leq 6$ along with $136$ random longer words, $16.3 \%$ of the test subwords (7,335 subwords) were unseen, and the model achieved $100 \%$ accuracy on the unseen data.




\section{Conclusions and future work}

Our experiments show that a small transformer can achieve high accuracy on combinatorial word-theoretic tasks in the infinite Coxeter group $\widetilde{A}_2$, and that its learned attention patterns can suggest mathematical structure. It would be interesting to apply these methods to other families of infinite Coxeter groups and to use more advanced mechanistic interpretability tools to extract and validate circuits implementing the known Brink-Howlett finite state automata \citet{Brink1993} for these groups. Note that implicit within the Brink-Howlett FSA is an algorithm to determine descent sets. It would be interesting to explicitly see this algorithm as computational circuits in the model.

\section{Generative AI statement} Generative AI helped with coding, figures, and literature search, but not with proofs or prose.


\appendix
\section{Group presentations and Coxeter groups}\label{app:GroupsandCoxeter}

The following background in group theory and the theory of Coxeter groups is standard. See for example \citet{artin2011algebra} and \citet{BjornerBrentiCoxeter}.

\subsection{Groups} 

A {\em group} $G$ is a set equipped with a binary operation \[\cdot: G \times G \rightarrow G\] called {\em multiplication} that satisfies the following properties:
\begin{enumerate}
        \item \emph{Associativity:} For all $a,b,c\in G$, $(a\cdot b)\cdot c=a\cdot(b\cdot c)$.
        \item \emph{Identity:} There exists an element $e\in G$, called the \emph{identity element}, such that $a\cdot e=e\cdot a=a$ for all $a\in G$.
        \item \emph{Inverses:} For every $a\in G$, there exists an element $a^{-1}\in G$, called the \emph{inverse} of $a$, such that $a\cdot a^{-1}=a^{-1}\cdot a=e.$
\end{enumerate}
A \emph{finite presentation} $G=\langle S\mid R\rangle$ of a group $G$ consists of: 
\begin{enumerate} 
\item A finite set $S$, whose elements are called the \emph{generators} of the presentation. 
\item A finite set $R$ of elements of the form $s_1^{n_1}\ldots s_l^{n_l}, n_l\in \mathbb{Z}, s_i\in S$, representing the identity element of $G$. Such elements are called the \emph{relations} of the presentation. 
\end{enumerate}
We assert that two words $w,w'$ represent the same group element if $w'$ can be obtained from $w$ by finitely many insertions or deletions of the relations.
\subsection{Coxeter groups} 

Let $S$ be an indexed finite set. A {\em Coxeter matrix} is a matrix given by the function,
\[
    m:S \times S \longrightarrow \mathbb{N} \cup \{\infty\}
\]
which satisfies
\begin{enumerate}
    \item $m(s,s') = m(s',s)  \textup{ for all } s,s' \in S$,
    \item $m(s,s') = 1 \text{ if and only if } s=s'$.
\end{enumerate}

The {\em Coxeter group} $W_m$ associated to Coxeter matrix $m$ is given by
$$
W_m = \langle S \mid (s_i s_j)^{m(s_i,s_j)}=e : s_i,s_j \in S \rangle.
$$
Here, the relations of the form $(s_i s_j)^{m(s_i,s_j)}=e$ for $i\neq j$ are called the \textit{braid relations}. On the other hand, since $m(s,s)=1$, every generator satisfies $s^2=e.$ Thus, two adjacent copies of the same generator in a given word may be removed without changing the represented group element.

\subsection{(Un)reduced words, minimal length words, and normal form} 
\label{sec:reduced-unreduced-words}
A \emph{word} with letters in $S$ is a finite sequence \[ w=s_1s_2\cdots s_k, s_i\in S. \] Its word length is the number of generators it contains: $|w|=k.$ In this paper, a word is called \emph{reduced} if it contains no two identical generators next to each other. Equivalently, $ s_i\neq s_{i+1} \text{ for all }1\leq i<k.$ 
A word is called \emph{unreduced} if it contains an adjacent repeated pair $ss$. 

A group element in a Coxeter group may be represented by many different words. The \emph{length} of an element $w\in W_m$, denoted $\ell(w)$ is given by \[ \ell(w) = \min \left\{ k: s'_1\cdots s'_k, s'_i\in S, w = s'_1\cdots s'_k \textup{ in } W_m \right\}. \] Likewise, we shall say a given representation $s_1\ldots s_k$ of $w$ is of \textit{minimal length} if $k=\ell(w)$. In general, such a minimal length representation is not unique, and so fixing an ordering $s_{i}<s_{i+1}$, we get reverse-lexicographically get a unique representative of $w$ among its minimal length representations, called the \textit{Reverse ShortLex normal form}. Alternatively, we shall call the unique expression of minimal lexicographic order the \textit{ShortLex normal form}.
\subsection{Descent sets} 
\label{sec:descent-sets}
Let $w\in W_m$. The \emph{right descent set} of $w$ is 
\[ 
D_R(w) = \left\{ s\in S: \ell(ws)<\ell(w) \right\}. 
\] 

Similarly, the \emph{left descent set} of $w$ is 
\[ 
D_L(w) = \left\{ s\in S: \ell(sw)<\ell(w) \right\}. 
\]

For every generator $s\in S$, multiplication by $s$ changes Coxeter length by exactly one: 
\[ 
\ell(ws)\in\{\ell(w)-1,\ell(w)+1\}, 
\]
and similarly, 
\[ 
\ell(sw)\in\{\ell(w)-1,\ell(w)+1\}. 
\] 
Likewise, $s\in D_R(w)$ if and only if some minimal length expression for $w$ ends in $s$, and $s\in D_L(w)$ if and only if some minimal length expression for $w$ begins with $s$.

\section{The Coxeter group $\widetilde{A}_2$}
\par 
\par Here, we prove Propositions \ref{prop:OddIndicesDetermineLeftDescent} and \ref{prop:6LetterEnough} through the proofs of Propositions \ref{prop:OddIndicesDetermineLeftDescentPrecise} and \ref{prop:6LetterEnoughPrecise}. We first see that Proposition \ref{prop:OddIndicesDetermineLeftDescent} is equivalent to:
\begin{proposition}\label{prop:OddIndicesDetermineLeftDescentPrecise}
    Let $w=s_1\ldots s_k$ be a word with letters $a_1, a_2, a_3$ of minimal length in $\widetilde{A}_2$. Then the left descent set of $w$ is $\{s_1, s_2\}$ if and only if there exists an odd index $i$ such that $s_i = s_{i+2}$. Otherwise $D_L(w)=\{s_1\}$.
\end{proposition}
This equivalence derives from the general Coxeter group fact that $D_L(w)=D_R(w^{-1})$, where $w^{-1}=s_k\ldots s_1$ for $w=s_1\ldots s_k$. Then, using this proposition, we prove Proposition \ref{prop:6LetterEnoughPrecise}, which implies Proposition \ref{prop:6LetterEnough}. The proofs also make use of algorithm detailed in \cite[Pseudocode, p. 13]{casselman2008computationcoxetergroupsii} for the left-multiplication of words in normal form by a generator. Foregoing a detailed discussion of the algorithm, we shall state here the minimal root look up table\cite[p.8]{casselman2008computationcoxetergroupsii} of $\widetilde{A}_2$:
\begin{center}
    \begin{tabular}{|c|c|c|c|}
        \hline
         & $a_1$ & $a_2$ & $a_3$ \\
        \hline
        $\alpha_{a_1}$ & $\ominus$ & $\alpha_4$ & $\alpha_6$ \\
        \hline
        $\alpha_{a_2}$ & $\alpha_4$ &$\ominus$ & $\alpha_5$\\
        \hline
        $\alpha_{a_3}$ & $\alpha_6$ & $\alpha_5$ &$\ominus$ \\
        \hline
        $\alpha_4$ & $\alpha_{a_2}$ & $\alpha_{a_1}$ & $\oplus$ \\
        \hline
        $\alpha_5$ & $\oplus$ & $\alpha_{a_3}$ & $\alpha_{a_2}$\\
        \hline
        $\alpha_6$ & $\alpha_{a_3}$ &$\oplus$ & $\alpha_{a_1}$\\
        \hline
    \end{tabular}
\end{center}
This table allows us to compute a certain value, $s\alpha$, where $s$ comes from one of the elements of the top row and $\alpha$ comes from one of the elements of the first column, which is an essential computation in the multipliation algorithm.\footnote{Explicitly, this computes the Weyl group action of the generators of $\widetilde{A}_2$ on its \textit{minimal roots}\cite[p. 7]{casselman2008computationcoxetergroupsii}. The notation $\ominus$ denotes the action of $s$ onto $\alpha_s$, and the notation $\oplus$ denotes $s\alpha$ producing a non-minimal positive root. That these are recorded in place of the roots is to provide sufficient conditions for the algorithm to end. In a particular, one understands that $\ell(sw)=\ell(w)-1$ if and only if $\ominus$ occurs somewhere in the root look up computation.}  Finally, throughout this section, for the sake of clarity and to prevent abuse of notation, we shall use $=$ to indicate equality on the level of words and $\sim$ to indicate equality on the level of group elements in $\widetilde{A}_2$. 

\begin{lemma}\label{lem:2Weyl1Root}
    Let $s_1, s_2, s_3$ be mutually distinct generators, and let $\alpha_{s}$ be as in \citet[p. 3]{casselman2008computationcoxetergroupsii} (that is the standard realization.) Then $s_2s_1\alpha_{s_2} = \alpha_{s_1}$
\end{lemma}
\begin{lemma}\label{lem:aba5LetterCase}
    Let $w = s_1\ldots s_5$ be a word with letters $a_1, a_2, a_3$ of minimal length in $\widetilde{A}_2$. If $s_3=s_5$ and $s_1 \neq s_3$ then $w\sim s_1s_2s_1s_3s_1$.
\end{lemma}
\begin{proof}
    As $s_1\neq s_3$, we see that $s_2$ must be distinct to $s_1$ and $s_4$. likewise $s_4=s_1$ since it cannot be $s_3$ and $s_2$: $s_3s_3\sim \textup{[empty]}$ and $s_2s_3s_2s_3 \sim s_3s_2$, which by the minimal length of $w$ implies that $w=s_1s_2s_3s_1s_3$. Then using the braid relation $s_1s_3s_1\sim s_3s_1s_3$, we get that $w\sim s_1s_2s_1s_3s_1$, as desired.
\end{proof}
\begin{lemma}\label{lem:abaGeneralCase}
    Let $w = s_1\ldots s_k$ be a minimal length word in the generators $a_1, a_2, a_3$ of  $\widetilde{A}_2$. If $s_i = s_{i+2}$ for some odd index $i$, then there exists a representation of $w$ starting with $s_1s_2s_1$. 
\end{lemma}
\begin{proof}
    If $s_1\ldots s_k$ is of minimal length, then $s_1\ldots s_l$ is of minimal length for all $l\leq k$. So the proof of this reduces to the case where $s_i=s_{i+2}$ if and only if $i=k-2$. In this case, we may proceed via recursion by Lemma \ref{lem:aba5LetterCase}, completing the proof. 
\end{proof}
\begin{proof}[Proof of Proposition \ref{prop:OddIndicesDetermineLeftDescentPrecise}]
    The if part of the proof follows from the lemmas above, and a direct calculation. Now to show the only if part, we see that there are two remaining cases to consider: $w=s_1\ldots s_k$ such that
    \begin{itemize}
        \item there is an even index $s_i = s_{i+2}$ but no odd index,
        \item there are no indices such that $s_i=s_{i+2}$.
    \end{itemize}
    \par Let us start by assuming that there is an even index $s_i = s_{i+2}$ but no such odd index. Then by an analogous argument to the one used in Lemma \ref{lem:abaGeneralCase}, there exists two minimal length representatives of $s_1\ldots s_k$ starting with $s_1s_2s_3s_2$ and $s_1s_3s_2s_3$ respectively. Now we recall (cf. \citet{casselman2008computationcoxetergroupsii}) that $D_L(w) = D_R(w^{-1})$ and that $s$ is a right descent of $w$ iff $w^{-1}$ applied to its associated simple root $\alpha_s$ evaluates negatively on the fundamental alcove (the purple triangle in Figure \ref{fig:A2tilde}). 
    
    We now consider the cases $\ell(w) := k=4$, and $k > 4$ separately. If $k=4$, we note first that by symmetry we may assume that $s_1 = a_1, s_2 = a_2, s_3 = a_3$. Then it is immediate from Figure \ref{fig:evenindexending} that $D_R(w^{-1}) = D_L(w) = \{s_1\}$, since its associated simple root $\alpha_{s_1}$ is the only one of the three that is negative on the fundamental alcove in the figure. If $k > 4$, then it is immediate from the fact that $D_R(s_1s_2s_3s_2) = \{s_2,s_3\}$ that $w \sim s_1s_2s_3s_2s_1w_0$ for some minimal length (possibly empty) word $w_0$. In this case, we need to argue that regardless of what $w_0$ is, $w^{-1} = (w_0)^{-1}(s_1s_2s_3s_2s_1)$ applied to $\alpha_{s_2}$ and $\alpha_{s_3}$ are both positive on the fundamental alcove, and hence neither $s_2$ nor $s_3$ is in $D_R(w^{-1}) = D_L(w)$. This can be seen by noting that since $s_1 \neq s_2 \neq s_3$, the action of $s_1s_2s_3s_2s_1$ on any alcove is to reflect it (and its markings by simple roots) across a hyperplane dominating $w_0^{-1}\alpha_{s_1}$, in the language of \citet{casselman2008computationcoxetergroupsii}. The conclusion follows from the observation that such a reflection will always send $w_0^{-1}(\alpha_{s_2})$ and $w_0^{-1}(\alpha_{s_3})$ to positive roots. See Figure \ref{fig:Dominatingreflection}.

    \begin{figure}
        \centering
        \includegraphics[width=0.60\linewidth]{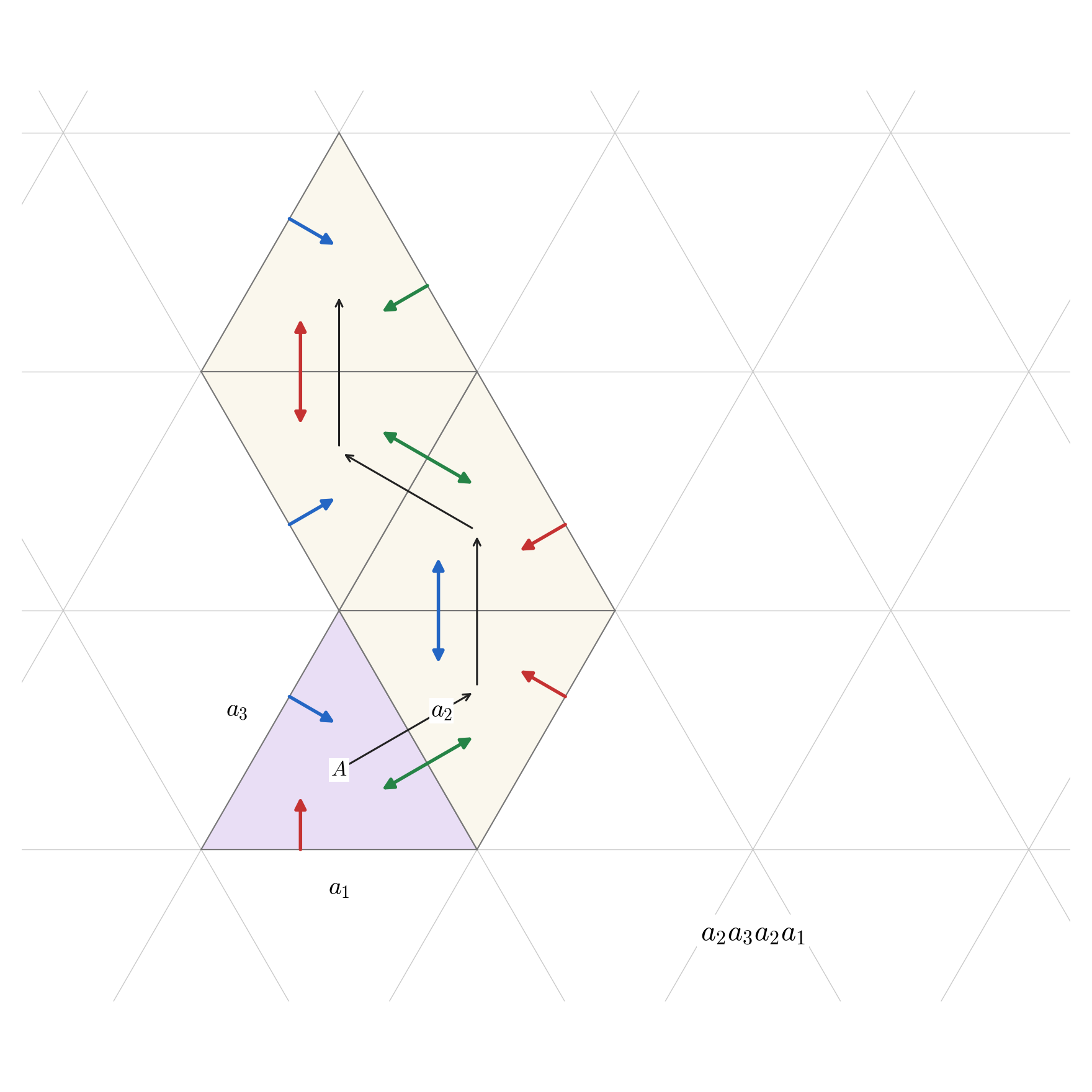}
        \caption{The gallery associated to $w^{-1} = a_2a_3a_2a_1$ for $w = a_1a_2a_3a_2$, with fundamental alcove $A$ (associated to the identity element of $\widetilde{A}_2$) marked in purple, and simple roots $\alpha_{a_1}, \alpha_{a_2}, \alpha_{a_3}$ marked in red, green, blue, respectively. Note that at the endpoint of the gallery, $w^{-1}(\alpha_{a_1})$ is negative, confirming that $a_1 \in D_L(w)$, and $w^{-1}(\alpha_{a_2}), w^{-1}(\alpha_{a_3}) $ are positive, confirming that $a_2,a_3 \not\in D_L(w)$.}
        \label{fig:evenindexending}
    \end{figure}

    \begin{figure}
        \centering
        \includegraphics[width=0.60\linewidth]{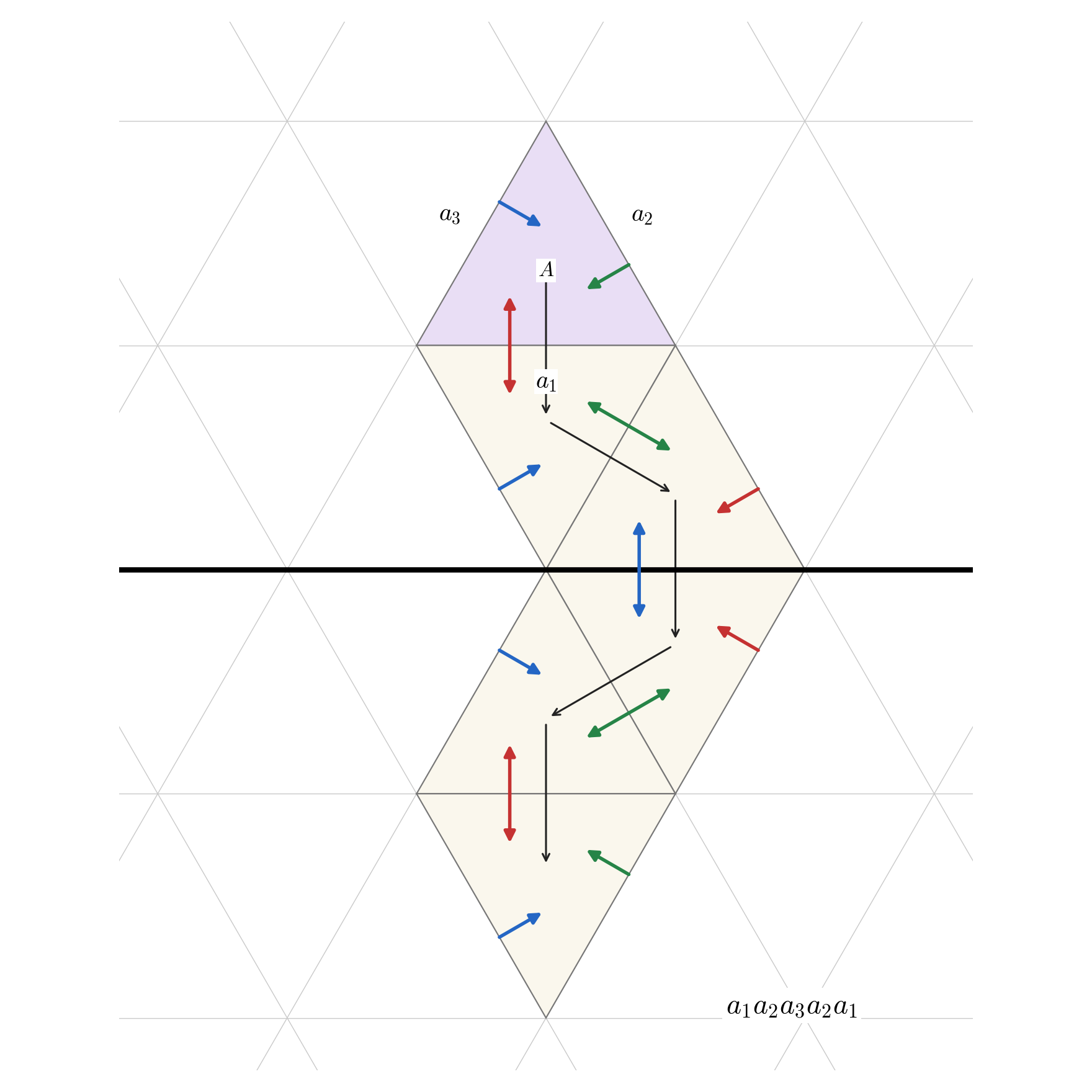}
        \caption{The gallery associated to $w^{-1} = a_1a_2a_3a_2a_1$ for $w=a_1a_2a_3a_2a_1$. Note that the markings on the alcove associated to the endpoint of this gallery are a reflected version of the initial alcove, along the hyperplane marked in bold, and both $w^{-1}(\alpha_{a_2}),w^{-1}(\alpha_{a_3})$ are therefore positive. If $w = a_1a_2a_3a_2a_1w_0$ for non-empty $w_0$, then the reflecting hyperplane will still dominate some minimal root in the fundamental alcove, and the conclusion will remain the same.}
        \label{fig:Dominatingreflection}
    \end{figure}
    \par Now to complete the proof let us assume there are no indices such that $s_i=s_{i+2}$. Then $w=(s_1s_2s_3)^jw_0$ where $w_0 = \textup{[empty]}, s_1, s_1s_2$. There are no braid relations in this word, and so its Reverse ShortLex normal form for any ordering is itself by Matsumoto's Theorem 
    \cite[Theorem 3.3.1]{BjornerBrentiCoxeter}. Taking $a_1<a_2<a_3$ and applying Lemma \ref{lem:2Weyl1Root} inductively we see that
    $$
    w^{-1}\alpha_{s_2} = \begin{cases}
        w^{-1}_0\alpha_{s_2} & 2\mid j\\
        w^{-1}_0\alpha_{s_3} & 2\nmid j
    \end{cases}.
    $$
    And so in all cases, $w^{-1}\alpha_{s_2}\neq \ominus$, which means that $\ell(s_2w) = \ell (w)+1$, giving us that it is not a left descent.
\end{proof}
\begin{proposition}\label{prop:6LetterEnoughPrecise}
    Let $w=s_1\ldots s_k$ be a word in ShortLex normal form with respect to the ordering $a_1<a_2<a_3$. Then the left descent set of $w$ is determined by the first 6 characters.
\end{proposition}
\begin{proof}
    To show this it suffices to show that the first six letters of a word in normal form determines whether it has an odd index $i$ with the property $s_i=s_{i+2}$. This follows immediately when either $s_1=s_3$ or $s_3=s_5$. So the only cases it remains to check are:
    \begin{itemize}
        \item $s_1s_2s_3s_4s_1s_4$,
        \item $s_1s_2s_3s_2s_1s_3$,
        \item $s_1s_2s_3s_1s_2s_3$.
    \end{itemize}
    where $s_i\neq s_{i+2}$ for all $i$ odd.
    \par Let us first consider $s_1s_2s_3s_4s_1s_4$. Let us suppose towards contradiction and say there exists some word $s_1s_2s_3s_4s_1s_4s_7\ldots s_k$ with odd $i\geq 7$ such that $s_{i}=s_{i+2}$. Then we see recursively by Lemma \ref{lem:abaGeneralCase} that
    $$
    s_1s_2s_3s_4s_1s_4s_7\ldots s_k \sim s_1s_2s_3s_4s_1s_4s_1\ldots
    $$
    which violates the minimality of $s_1s_2s_3s_4s_1s_4s_7\ldots s_k$ as $s_4s_1s_4s_1=s_1s_4$. So no such odd index can exist. We argue similarly for the cases $s_1s_2s_3s_2s_1s_3$. 
    \par Finally in the case $s_1s_2s_3s_1s_2s_3$, we likewise see that any $s_1s_2s_3s_1s_2s_3s_7\ldots s_k$ with odd $i\geq 7$ such that $s_{i}=s_{i+2}$ propagates back via the braid relations to a violation of lexicographic minimality.
\end{proof}

\section{Masking experiments}
\begin{figure}[h]
    \centering
    \begin{subfigure}{195 pt}
    \centering
    \includegraphics[width=\linewidth]{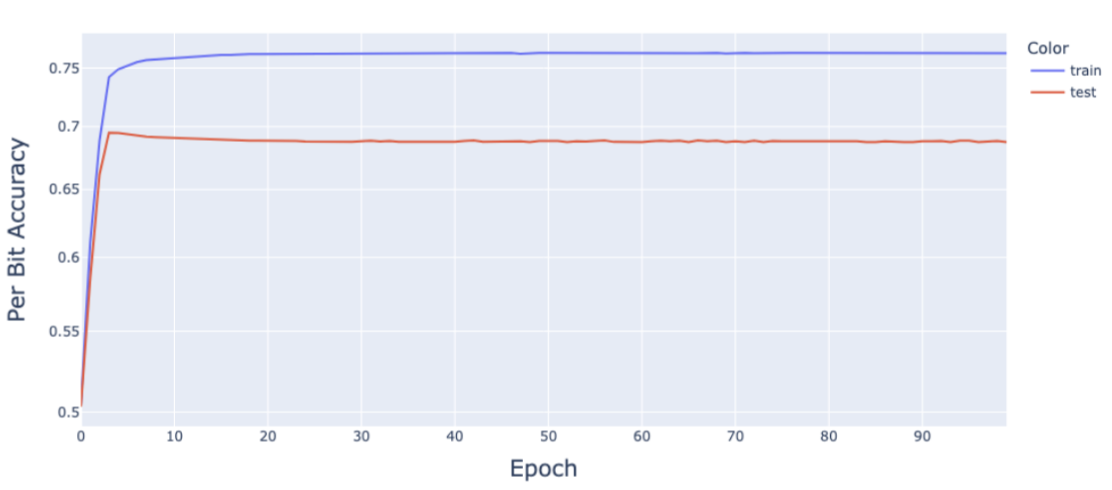}
    \caption{First $5$ tokens masked}  \label{fig:First5MaskedAccuracy}    
    \end{subfigure} \hfill
    \begin{subfigure}{195 pt}
    \centering    \includegraphics[width=\linewidth]{MaskAfterFirst5.png}
    \caption{All but first $5$ tokens masked}  \label{fig:AllButFirst5MaskedAccuracy}
\end{subfigure}
\caption{Model accuracy for the left descent task on ShortLex words in $\widetilde{A}_2$ when a subset of input tokens is masked during training. Test and train accuracy were nearly $100\%$ on the left descent task when no tokens were masked, stayed at nearly $100\%$ when all but the first $5$ tokens were masked, and dropped to below $70\%$ on the test set when the first $5$ tokens were masked.}
\label{fig:MaskingAccuracy}
\end{figure}

To test whether the information contained in the first few tokens is sufficient for predicting the left descent set of a word of length \(\leq 36\), we ran two token-masking experiments on the ShortLex data set: one in which we masked the first \(5\) tokens, and one in which we masked all but the first \(5\). Consistent with the attention pattern in Figure \ref{fig:LeftDescents}, performance dropped substantially when the first \(5\) tokens were masked, while remaining nearly unchanged when only the first \(5\) tokens were retained. See Figure \ref{fig:MaskingAccuracy}.

\section{Model training details and attention pattern consistency}
\emph{Model and training.} We used the TransformerLens \texttt{HookedTransformer}: $1$ layer, $d_{\mathrm{model}}=256$, $4$ heads of dimension $64$, MLP width $1024$ with ReLU, no layer normalization, learned absolute positional embeddings, causal attention, vocabulary of $3$ generators plus padding, and $3$ sigmoid outputs. All biases were frozen at zero for interpretability. There are $799{,}747$ parameters in total. Optimizer: AdamW, learning rate $10^{-4}$ (constant), $\beta=(0.9,0.98)$, weight decay $0.5$, full batch, masked binary cross-entropy. Epochs: $10{,}000$ (left) and $25{,}000$ (right). Seeds: split seed $0$, initialization seed $999$, and training seed $598$ for the published runs. Replicates $s=1,2,3$ use split seed $s$ and initialization and training seeds offset by $s$. 

\emph{Consistency of the attention patterns over multiple runs}. Test exact-set accuracy over the four runs for the ShortLex data set is $1.0000\pm0.0000$ (left descent task) and $0.9996\pm0.0004$ (right descent task), with train accuracy $1.0000$ throughout. The left descent attention concentration on the first six tokens recurred with every seed ($44$--$51\%$ of final-token attention on keys $1$--$6$, vs.\ $17\%$ uniform). The right descent effect (same parity keys matter most) also recurred in every seed.

\section{Random and reduced word experiments}

\begin{figure}[h]
    \centering
    \begin{subfigure}{195 pt}
    \centering
    \includegraphics[width=\linewidth]{random_length_10_acc.png}
    \caption{Random words of length $10$}  \label{fig:RandomAccuracy10}    
    \end{subfigure} \hfill
    \begin{subfigure}{195 pt}
    \centering    \includegraphics[width=\linewidth]{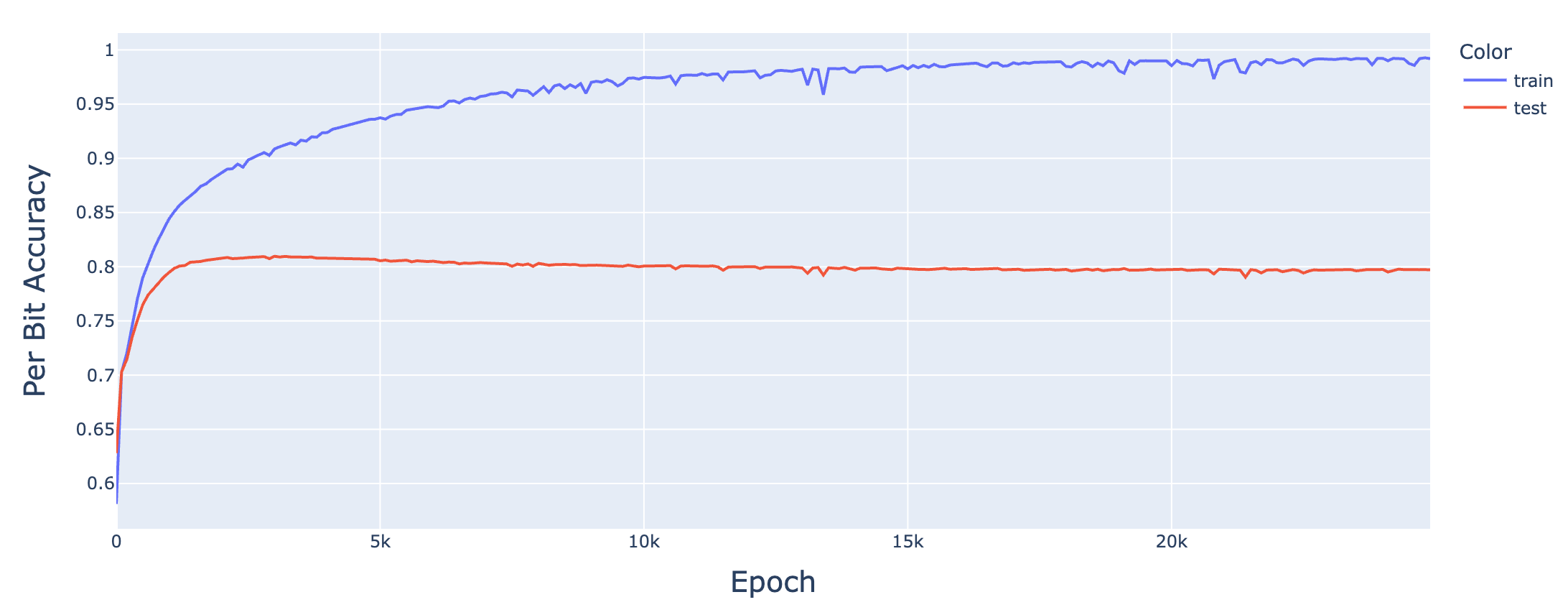}
    \caption{Random words of length $16$}  \label{fig:RandomAccuracy16}
\end{subfigure}
\caption{Model accuracy for the left descent task on random words in $\widetilde{A}_2$.}
\label{fig:LeftRandomAccuracy}
\end{figure}

\begin{figure}[h]
    \centering
    \begin{subfigure}{195 pt}
    \centering
    \includegraphics[width=\linewidth]{reduced_length_10_acc.png}
    \caption{Reduced words of length $10$}  \label{fig:ReducedAccuracy10}    
    \end{subfigure} \hfill
    \begin{subfigure}{195 pt}
    \centering    \includegraphics[width=\linewidth]{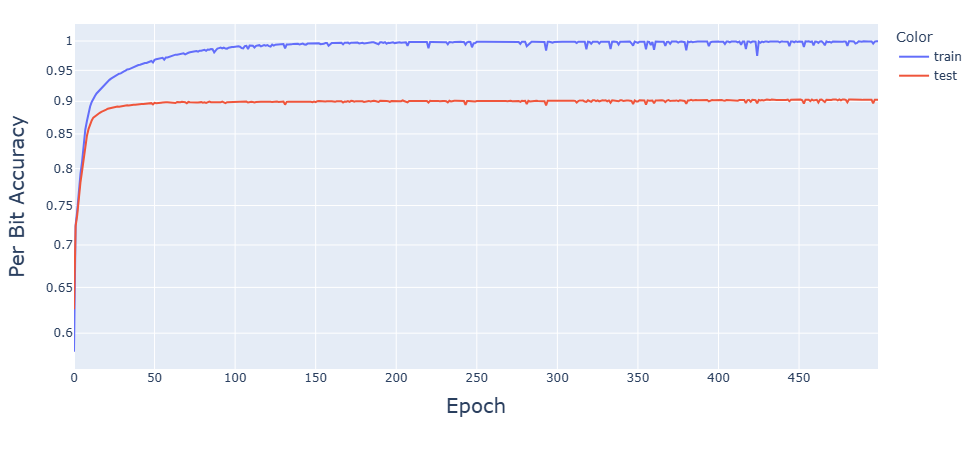}
    \caption{Reduced words of length $16$}  \label{fig:ReducedAccuracy16}
\end{subfigure}
\caption{Model accuracy for the left descent task on reduced words in $\widetilde{A}_2$}
\label{fig:LeftReducedAccuracy}
\end{figure}

The work reported in the main body of the paper came out of training on labeled data sets of words in ShortLex normal form in the Coxeter group $\widetilde{A}_2$. We also trained the same transformer model on the less-structured data sets of random and reduced words, where random words have letters sampled uniformly from $\{a_1, a_2, a_3\},$ while reduced words are sampled uniformly as above and then {\em reduced} by eliminating all (redundant) substrings of the form $a_ia_i$. We trained on datasets of $2000$ words at lengths $10$ and $16$. Figures \ref{fig:LeftRandomAccuracy} and \ref{fig:LeftReducedAccuracy} record the train/test accuracy of the model on the left descent task for all $4$ data sets after training has stabilized.

\begin{figure}[h]
    \centering
    \begin{subfigure}{190 pt}
    \centering
    \includegraphics[width=\linewidth]{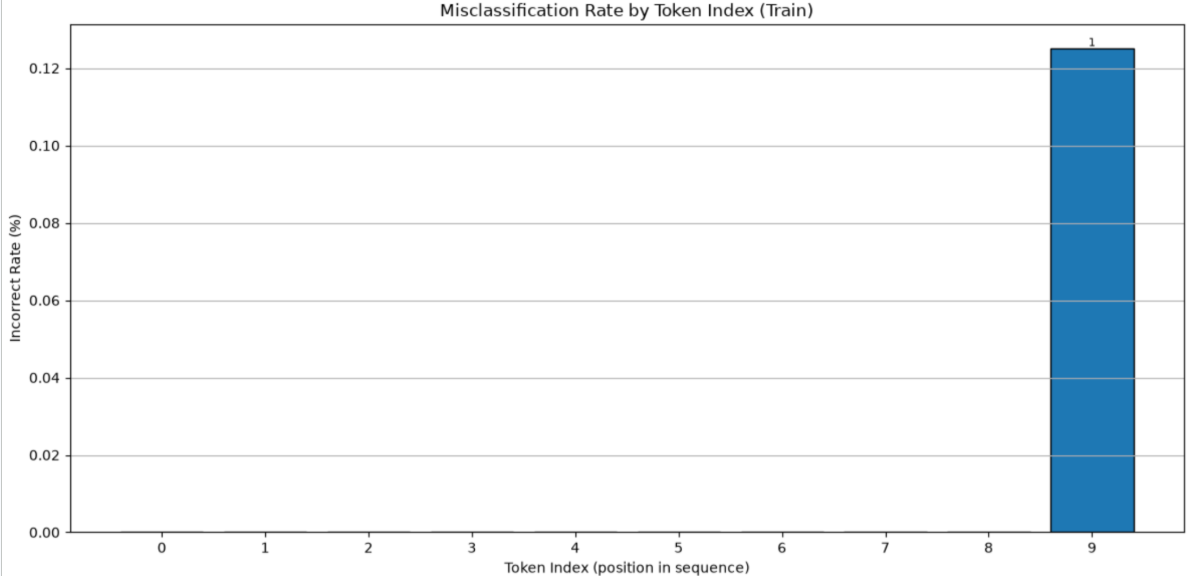}
    \caption{Training set misclassification by index.}  \label{fig:RandomTrainMisclass}    
    \end{subfigure} \hfill
    \begin{subfigure}{190 pt}
    \centering    \includegraphics[width=\linewidth]{test_misclass_random.png}
    \caption{Test set misclassification by index.}  \label{fig:RandomTestMisclass}
\end{subfigure}
\caption{Misclassification rates by index for random length $10$ words in $\widetilde{A}_2$}
\label{fig:MiscassificationRandom}
\end{figure}

\begin{figure}[h]
    \centering
    \begin{subfigure}{190 pt}
    \centering
    \includegraphics[width=\linewidth]{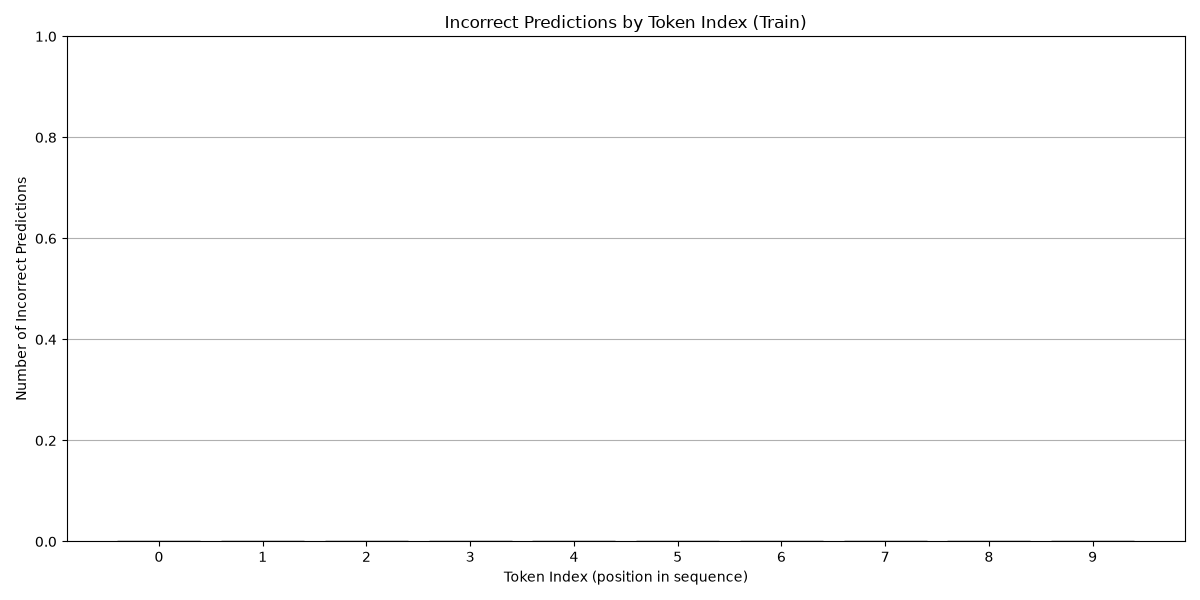}
    \caption{Train misclassifications for left descent of reduced length 10 words in $\widetilde{A}_2$.}  \label{fig:ReducedTrainMisclass}    
    \end{subfigure} \hfill
    \begin{subfigure}{190 pt}
    \centering    \includegraphics[width=\linewidth]{test_misclass_reduced.png}
    \caption{Test misclassifications for left descent of reduced length 10 words in $\widetilde{A}_2$.}  \label{fig:ReducedTestMisclass}
\end{subfigure}
\caption{Misclassification rates by index for reduced length $10$ words in $\widetilde{A}_2$}
\label{fig:MisclassificationReduced}
\end{figure}

Figures \ref{fig:MiscassificationRandom} and \ref{fig:MisclassificationReduced} report misclassification rate by index on the train and test sets for random and reduced words of length $10$. Observe that the misclassification rate grows rapidly with the index, suggesting that the model has not learned a length-generalizable algorithm for determining the left descent set, in contrast to the performance of the model on the ShortLex data set.

\subsubsection*{Acknowledgements}
The authors are indebted to Tony Licata and Joel Gibson, who generously shared expertise and references on computation in Coxeter groups. The descent set task was Tony's idea. We gratefully acknowledge the support of the Institute for the Foundations of Machine Learning (IFML), which funded P. Ramanakrishnan's project to adapt the algorithm outlined in \citet{casselman2008computationcoxetergroupsii} to produce the data sets used to train our models. Y. Huang, A. Katyal, R. Lin, and W. Wang were supported by Boston College Undergraduate Research Fellowships to do the work on this  project during Summer 2026 through the Boston College Experimental Math and Machine Learning Lab (BCExpMMLL). This project built on a previous BCExpMMLL project with Caleb Ashley, David Ruiz De Castilla, Vanessa Rigoglioso, Roman Roy, Rob Vollbrecht, and Celina Xiong. 

\bibliographystyle{plainnat} 
\bibliography{neurips_2026}    
\end{document}